\documentclass[10pt,twocolumn,letterpaper]{article}

\usepackage{cvpr}              

\usepackage{graphicx}
\usepackage{amsmath}
\usepackage{amssymb}
\usepackage{booktabs}
\usepackage{amsthm}
\usepackage{rotating}
\usepackage{float}

\usepackage{MnSymbol,bbding,pifont}

\usepackage{multirow}
\usepackage[table,xcdraw]{xcolor}

\newtheorem{proper}{Property}

\usepackage{enumitem}
\usepackage{float}

\usepackage{fontawesome5}
\makeatletter
\@namedef{ver@everyshi.sty}{}
\makeatother
\usepackage{tikz}
\usepackage{pifont}
\newcommand{\xmark}{\ding{55}}%
\newcommand{\cmark}{\ding{51}}%

\usepackage{color}

\usepackage[table,xcdraw]{xcolor}

\definecolor{olive}{rgb}{0.6, 0.6, 0.2}
\definecolor{sand}{rgb}{0.8666666666666667, 0.8, 0.4666666666666667}
\definecolor{wine}{rgb}{0.5333333333333333, 0.13333333333333333, 0.3333333333333333}

\definecolor{deblue}{RGB}{11,132,147}
\definecolor{ocra}{RGB}{204, 119, 34}

\newcommand{\fcircle}[2][red,fill=red]{\tikz[baseline=-0.5ex]\draw[#1,radius=#2] (0,0.03) circle ;}

\newcommand{\cUnet}{\textcolor{black}{{\fontfamily{cmtt}\selectfont continuous U-Net~}}}
\newcommand{\cUnets}{\textcolor{black}{{\fontfamily{cmtt}\selectfont continuous U-Nets~}}}
\newcommand{\CUnet}{\textcolor{black}{{\fontfamily{cmtt}\selectfont Continuous U-Net~}}}

\definecolor{electricindigo}{rgb}{0.44, 0.0, 1.0}
\definecolor{lightblue}{RGB}{240,245,255}
\definecolor{darkblue}{RGB}{40,40,85}
\definecolor{babyblue}{rgb}{0.54, 0.81, 0.94}
\definecolor{pearDark}{HTML}{2980B9}
\definecolor{pearDarker}{HTML}{1D2DEC}

\usepackage[pagebackref,breaklinks,colorlinks]{hyperref}
\usepackage{multirow}

\usepackage[capitalize]{cleveref}
\crefname{section}{Sec.}{Secs.}
\Crefname{section}{Section}{Sections}
\Crefname{table}{Table}{Tables}
\crefname{table}{Tab.}{Tabs.}

\definecolor{indigo(web)}{rgb}{0.29, 0.0, 0.51}
\usepackage[framemethod=TikZ]{mdframed}

\newcounter{theob}[section] 
\renewcommand{\thetheob}{\arabic{theob}}
\newenvironment{theob}[2][]{%
\refstepcounter{theob}%
\ifstrempty{#1}%
{\mdfsetup{%
frametitle={%
\tikz[baseline=(current bounding box.east),outer sep=0pt]
\node[anchor=east,rectangle,fill=blue!20]
{\strut Theorem~\thetheob};}}
}%
{\mdfsetup{%
frametitle={%
\tikz[baseline=(current bounding box.east),outer sep=0pt]
\node[anchor=east,rectangle,fill=blue!20]
{\strut Theorem~\thetheo:~#1};}}%
}%
\mdfsetup{innertopmargin=1pt,linecolor=blue!20,%
linewidth=2pt,topline=true,%
frametitleaboveskip=\dimexpr-\ht\strutbox\relax
}
\begin{mdframed}[]\relax%
\label{#2}}{\end{mdframed}}

\newcounter{lem}[section] 
\renewcommand{\thelem}{\arabic{lem}}

\newcounter{prf}[section]
\newenvironment{prf}[2][]{%
\refstepcounter{prf}%
\ifstrempty{#1}%
{\mdfsetup{%
frametitle={%
\tikz[baseline=(current bounding box.east),outer sep=0pt]
\node[anchor=east,rectangle,fill=black!10]
{\strut Global Error};}}
}%
{\mdfsetup{%
frametitle={%
\tikz[baseline=(current bounding box.east),outer sep=0pt]
\node[anchor=east,rectangle,fill=black!10]
{\strut Proof~\theprf:~#1};}}%
}%
\mdfsetup{innertopmargin=1pt,linecolor=black!20,%
linewidth=2pt,topline=true,%
frametitleaboveskip=\dimexpr-\ht\strutbox\relax
}
\begin{mdframed}[]\relax%
\label{#2}}{\end{mdframed}}
\newcounter{propb}[section] 
\renewcommand{\thepropb}{\arabic{propb}}
\newenvironment{propb}[2][]{%
\refstepcounter{propb}%
\ifstrempty{#1}%
{\mdfsetup{%
frametitle={%
\tikz[baseline=(current bounding box.east),outer sep=0pt]
\node[anchor=east,rectangle,fill=green!20]
{\strut Proposition~\thepropb};}}
}%
{\mdfsetup{%
frametitle={%
\tikz[baseline=(current bounding box.east),outer sep=0pt]
\node[anchor=east,rectangle,fill=green!20]
{\strut Proposition~\thepropb:~#1};}}%
}%
\mdfsetup{innertopmargin=1pt,linecolor=green!20,%
linewidth=2pt,topline=true,%
frametitleaboveskip=\dimexpr-\ht\strutbox\relax
}
\begin{mdframed}[]\relax%
\label{#2}}{\end{mdframed}}

\def\cvprPaperID{10002} 
\def\confName{CVPR}
\def\confYear{2022}

\begin{document}

\title{Continuous U-Net: \\Faster, Greater and Noiseless}

\author{Chun-Wun Cheng$^{1,2\ast\ddagger}$, Christina Runkel$^{2\ast}$, Lihao Liu$^{2}$\thanks{Equal Contribution. $\ddagger$Work done whilst interning in Cambridge. }, Raymond H Chan$^{1}$, \\ Carola-Bibiane Schönlieb$^{2}$, Angelica I Aviles-Rivero$^{2}$
\\  \:
$^{1}$ City University of Hong Kong (CityU), Hong Kong  \\ $^{2}$ University of Cambridge, UK 
}
\maketitle

\begin{abstract}
Image segmentation is a fundamental task in image analysis and clinical practice. The current state-of-the-art techniques are based on U-shape type encoder-decoder networks with skip connections, called U-Net. Despite the powerful performance reported by existing U-Net type networks, they suffer from several major limitations. Issues include the hard coding of the receptive field size, compromising the performance and computational cost, as well as the fact that they do not account for inherent noise in the data. They have problems associated with discrete layers, and do not offer any theoretical underpinning. In this work we introduce  continuous U-Net, a novel family of networks for image segmentation. 
Firstly, continuous U-Net is a continuous deep neural network that introduces new dynamic blocks modelled by second order ordinary differential equations. Secondly, we provide theoretical guarantees for our network demonstrating faster convergence, higher robustness and less sensitivity to noise.
Thirdly, we derive qualitative measures to tailor-made segmentation tasks.
We demonstrate, through extensive numerical and visual results, that our model
outperforms existing U-Net blocks for several medical image segmentation benchmarking datasets. 
\end{abstract}
\section{Introduction}





\begin{figure}
    \centering
    \includegraphics[width=1\linewidth]{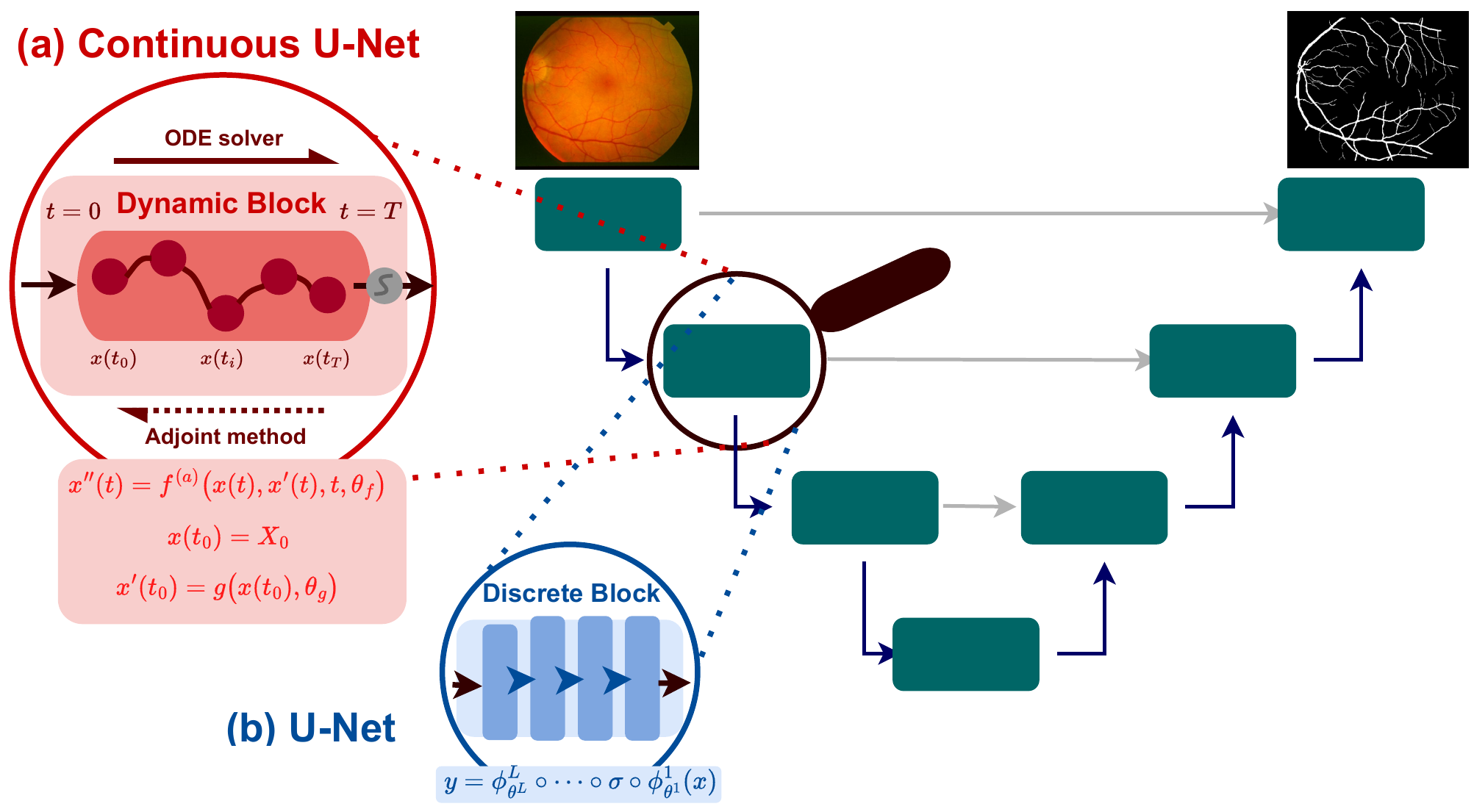}
    \caption{Visual comparison of our \cUnet vs. U-Net (and variants). The zoom-in views display the difference between discrete blocks in U-Net  and our proposed dynamic blocks.    
    }
    \label{fig:teaser}
\end{figure}
Image segmentation is a fundamental task in medical image analysis and clinical practice. It is a critical component in several  applications including diagnosis, surgery-guided planning and therapy. Manual segmentation of such medical datasets is time-consuming, and increases financial cost. The advent of deep learning, and in particular Fully Convolutional Neural Networks (FCNNs)~\cite{long2015fully}, opened the door to automatic segmentation techniques. The current state-of-the-art techniques are based on U-shape type  encoder-decoder networks with skip connections~\cite{qin2020u2,liu2020psi,zhou2019unet++,su2021msu,valanarasu2022unext}. This family of networks has demonstrated astonishing results due to their representation learning capabilities, and the ability to recover fine-grained details.

The seminal paper of Ronneberger et al.~\cite{ronneberger2015u} introduced the U-Net model
for Biomedical Image Segmentation. More precisely,  U-net seeks to capture both the context and the localisation features. Firstly, it uses skip connections to provide additional information that helps the decoder to generate better semantic features. Secondly, it consists of a symmetric encoder-decoder scheme, which reduces the computational cost. The impressive performance of U-Net motivated the fast development of several U-Net variants e.g.,~\cite{qin2020u2,su2021msu,valanarasu2022unext, etmann2020iunets}, and its usage in wide variety of clinical data. 


Despite the powerful performance reported by existing U-Net type networks, \textit{they suffer from several major limitations.} \fcircle[fill=wine]{2pt}  Firstly, they hard code the receptive field size, which requires architecture optimisation for each segmentation task. Wider receptive views  increase the accuracy~\cite{chen2018encoder}. However, the limited computational memory forces a trade-off between network depth, width, batch size, and input image size. Since deep learning makes the prediction by discretising the solution layer by layer, this has a very high computation cost.  \fcircle[fill=deblue]{2pt} Secondly, the current existing U-Net models do not account for inherent noise that affects the predictions. \fcircle[fill=ocra]{2pt} Thirdly, U-Net is a discrete neural network with discrete layers, but medical image data is continuous.  \fcircle[fill=olive]{2pt} Fourthly, existing U-net variants do not provide any theoretical underpinning. 
%
%
%
%

\textit{Can one design a U-type network that overcomes the aforementioned major issues of existing models?} This is the question that we address in this work. Our work is motivated by deep implicit learning, and continuous approaches based on Neural Ordinary Differential Equations (Neural ODEs)~\cite{chen2018neural,dupont2019augmented} in particular. The body of literature has explored Neural ODEs mainly for image classification tasks. 
So far, Neural ODEs have only been used for a small number of applications as designing Neural ODEs for more complex tasks is far from being trivial. 


We underline that U-Nets (and majority of its variants) are designed as discrete neural networks.This is in contrast to the continuous nature of medical data. In this regard, Recurrent Neural Networks (RNNs) are more appropriate. However, non-uniform intervals for handling medical data can compromise the performance. Neural ODEs are suitable for continuous data since they output a vector field. Another major problem of the discrete neural network is the issue of overfitting/underfitting  \cite{bilbao2017overfitting}. However, we will show that looking at U-Net from the lens of continuous dynamical systems, this issue can be mitigated.


Another major keypoint is that the performance of image segmentation depends partly on the receptive field of the networks. Due to the limited GPU memory, U-Net and other variants force a trade-off between architecture design and input image size.
Different techniques have been developed for this problem. Examples are dilated convolutions~\cite{folle2019dilated} and reversible blocks~\cite{brugger2019partially}. These architectures reduce the computational cost because of the reduced number of stored active functions.  However, the memory cost of these architectures is still associated with the depth of the model. In our work, we seek to address this problem using 
a trace operation via the adjoint sensitivity method with $\mathcal{O}(1)$ memory cost. That is, for neural networks based on ODEs, reversibility is built into the architecture by construction.
Therefore, no matter the model's complexity, one can always provide the benefit of constant memory cost.
%
In this work, we propose a novel family of networks that we call \cUnets for medical image segmentation. An overview of our approach in contrast to a standard U-Net can be found in Figure \ref{fig:teaser}. In particular, our contributions are:


\smallskip

\faHandPointRight[regular]  We propose \cUnet a new network for medical image segmentation. \CUnet is a continuous deep network whose dynamics are modelled by second order ordinary differential equations.  We view the dynamics in our network as the boxes consisting of CNNs and transform them into dynamic blocks to get a solution. We introduce the first 
U-Net variant working explicitly in higher-order neural ODEs. We highlight:
    \begin{itemize} [noitemsep,nolistsep]
        \item \textbf{Faster Convergent.} By modelling the dynamics in a higher dimension, we provide more flexibility to learn the trajectories. Therefore, \cUnet requires fewer iterations for the solution, which is more computationally efficient and in particular provides constant memory cost {(Proposition~\ref{prop:higher_order}  \& Theorem~\ref{cor:adjoint}).}
        \item \textbf{Greater: More Robustness.} We show that \cUnet is more robust than other variants (CNNs) and provide an  intuition for this (Theorem~\ref{theo:robustness}).  Moreover, we demonstrate that our dynamic blocks are reliably useful (Theorem~\ref{theo:well-posed}).
        \item \textbf{Noiseless.}  We show that \cUnet is always bounded by some range whilst CNNs are not. Also, our network is smoother than existing ones leading to  better handling the inherent noise in the data. 
        \item \textbf{Underpinning Theory.} \CUnet is the first U-type network that comes with theoretical understanding. 
    \end{itemize}
    
 \faHandPointRight[regular] \textbf{Open the ``box'' of ODE-solvers.} Existing works lack guidelines for choosing the best ODE solver. We derive qualitative measures for the choice of different ODE-solvers.  At the practical level, this means that our framework can be tailor-made for various segmentation tasks (Global Error \& Theorem~\ref{thm:GE}).

\faHandPointRight[regular] We demonstrate, through extensive numerical and visual results, that our proposed \cUnet outperforms existing U-type blocks. Moreover, we show that our proposed network stands alone performance-wise without any additional mechanism, and its performance readily competes with other mechanisms including transformers, nested U-Nets and tokenised MLP.


\section{Related Work}
The task of image segmentation, for medical data, has been widely investigated in the community, whith state-of-the-art techniques relying in U-type architectures, alone or in combination with additional mechanisms. In this section, we review the existing techniques in turn.

\subsection{U-type Nets: A Block Based Perspective}
The gold-standard U-Net model was introduced by Ronneberger et al.~\cite{ronneberger2015u} for biomedical image segmentation. U-Net consists of four main components -- blocks of neural network layers, downsampling, upsampling  and concatenation operations. The astonishing performance reported by U-Net motivated the fast development of a wide range of U-type variants, where the key difference is largely based on different types of blocks.  

The standard U-Net architecture~\cite{ronneberger2015u} uses convolutional blocks, followed by an activation function. Further work in this area was introduced in~\cite{zhang2018road}, where the authors use residual blocks~\cite{he2016deep} yielding to ResUNet. To facilitate learning, the input $x$ of each convolutional block $F_{\theta}$ is added to the output via skip-connections so that $y~=~F_{\theta}(x) + x$.  Li et al. introduced DenseUNet~\cite{li2018h} where densely connected blocks are used in the U-Net structure. Their work follows the principles of DenseNet~\cite{huang2017densely}. A different U-Net variant is based on inception blocks~\cite{zhang2018bi}. The notion is based on computing convolutions with varying kernel sizes in parallel to then concatenate their outputs. Common kernel sizes are $1\times 1$, $3\times 3$ and $5\times 5$. This approach aims at computing different levels of features by choosing different kernel sizes. 

Another U-type network uses  pyramid pooling blocks~\cite{zhao2017pyramid}, where the idea is to pool the input to get different input sizes (e.g., $1\times 1 \times C$, $2 \times 2 \times C$ and $4 \times 4 \times C$ for the number of channels $C$), and apply convolutions on each of them  before upsampling to the original input size. The upsampled output is then concatenated channel-wise. Most recently, the work of Pinckaers et al.~\cite{pinckaers2019neural} uses first order neural ODEs in a U-Net setting. 

\subsection{U-type Nets with Additional Mechanisms } 
The aforementioned techniques seek to create variants of U-Net solely based on the block types. Another set of techniques has instead focused on creating additional mechanism in U-Net structures. Zhang et al. introduced ResUNet~\cite{zhang2018road}, where residual blocks along with several additional mechanisms such as atrous convolutions \cite{chen2017rethinking}, pyramid scene parse pooling \cite{zhao2017pyramid} and multi-task inference \cite{ruder2017overview} were used.  An attention mechanism was introduced in U-Net where the combination of both is known as Attention U-Net~\cite{oktay2018attention}. Attention gates are used to filter features before concatenating the upsampled input and the skip connection in the decoder part of the U-Net. 

In more recent works, DynUNet~\cite{ranzini2021monaifbs} was introduced as a combination of two works. It takes the heuristic rules and setting from  nnU-Net \cite{isensee2019automated}, and the optimisation scheme of Futrega et al.~\cite{futrega2022optimized} for searching an (sub-)optimal network structure. Self-attention mechanisms have also been explored for U-Net. Chen et al. proposed TransUNet~\cite{chen2021transunet} that combines  a transformer encoder, with a self-attention mechanism, and a classical convolutional neural network decoder. The most recent U-type network, called UNeXt, was introduced in~\cite{valanarasu2022unext}.  UNeXt is a convolutional multi-layer perceptron (MLP) based network, which uses tokenised MLP blocks with axial shifts. In comparison to transformer based approaches like TransUNet, UNeXt needs only a small number of parameters.

\begin{table}[t!] 
\caption{Overview of properties of our \cUnet vs. existing U-type networks.}
\centering
\resizebox{0.47\textwidth}{!}{
\begin{tabular}{ccc}
\hline
\rowcolor[HTML]{EFEFEF} 
\textsc{Properties} &  \cUnet  & \textsc{Other U-Nets} \\ \hline
\multicolumn{1}{c|}{\fcircle[fill=deblue]{2pt} Parameters Efficiency} & \textcolor{blue}{\cmark}  & \textcolor{red}{\xmark} \\
\multicolumn{1}{c|}{\fcircle[fill=deblue]{2pt} Constant Memory Cost} &  \textcolor{blue}{\cmark}  &  \textcolor{red}{\xmark}\\
\multicolumn{1}{c|}{\fcircle[fill=deblue]{2pt} Continuous Network} &  \textcolor{blue}{\cmark}  & \textcolor{red}{\xmark} \\
\multicolumn{1}{c|}{\fcircle[fill=deblue]{2pt} Noise Resistant} &  \textcolor{blue}{\cmark}  & \textcolor{red}{\xmark}\\
\multicolumn{1}{c|}{\begin{tabular}[c]{@{}c@{}}\fcircle[fill=deblue]{2pt} Beyond homomorphic\\ Transformations\end{tabular}} &  \textcolor{blue}{\cmark}  &  \textcolor{red}{\xmark}\\
\multicolumn{1}{c|}{\fcircle[fill=deblue]{2pt} Reversible} &   \textcolor{blue}{\cmark} &  \textcolor{red}{\xmark}\\
\multicolumn{1}{c|}{\fcircle[fill=deblue]{2pt} Theoretical Underpinning} &   \textcolor{blue}{\cmark}  & \textcolor{red}{\xmark}\\ \hline
\end{tabular}
}
\label{table:properties}
\end{table}


\subsection{Existing Techniques \& Comparison to Ours}
We provide a summary of properties for \cUnet and existing U-Type nets in Table~\ref{table:properties}. More precisely, existing U-type networks~\cite{ronneberger2015u,zhang2018road,li2018h,zhang2018road,oktay2018attention,ranzini2021monaifbs,valanarasu2022unext} discretise the solution layer by layer which yields to high computational cost. Contrary, our work is a continuous architecture that can be solved via the adjoint method\cite{chen2018neural}, which translates in constant memory cost. Unlike existing U-type blocks~\cite{ronneberger2015u,zhang2018road,he2016deep,li2018h,zhang2018bi,zhao2017pyramid}, we propose new dynamic blocks modelled by second-order neural ODEs, which are not restricted to homomorphic transformations. Moreover, opposite to~\cite{pinckaers2019neural} 
our dynamic blocks are at least twice continuously differentiable resulting in being robust to noise.

Most recently, new mechanisms have been used along with U-Net including attention, self-attention, transformers, heuristic rules and nested U-Nets~\cite{zhang2018road, oktay2018attention, ranzini2021monaifbs,chen2021transunet,valanarasu2022unext}. Unlike these works, our \cUnet does not use any additional mechanism. We instead introduce new dynamic blocks. Therefore, the philosophy of our work is closer to the block based U-Net techniques~\cite{ronneberger2015u,zhang2018road,he2016deep,li2018h,zhang2018bi,zhao2017pyramid} and not directly comparable to those with additional mechanisms. We highlight that our \cUnet does not use any additional mechanism, \textit{opening the door to a new research line on continuous U-type networks.}
Finally and to the best of our knowledge, this is the first U-type architecture that provides underpinning theory.

\section{Proposed Technique}
This section contains the key parts of our proposed \cUnet: (i) we present our proposed dynamic blocks and
how we model the dynamics of our network using second order Neural ODEs, (ii) our derived quality measure for tailor-made segmentation tasks and (iii) robustness and noise properties of our network.

\subsection{Unboxing Continuous U-net}
\CUnet greatly differs from existing U-type networks since it is modelled  as a continuous approach. That is, we avoid computing predictions by discretising the solution layer-by-layer, which involves a high computational cost. In particular, our network models the dynamics by taking the CNN boxes and transforming them into ODE blocks. Unlike existing works on Neural ODEs, we go beyond the standard learning setting by designing a new U-type architecture using higher order neural ODEs. 
\textit{Why designing higher-order blocks?} The standard Neural ODE setting fails to learn complex flows. As the training progresses and the flow becomes more complex,  the number of steps required to solve the ODE increases \cite{chen2018neural,grathwohl2018ffjord}. This is one of the limitation of neural ODEs. Although augmented neural ODEs (ANODE)~\cite{dupont2019augmented} were proposed to mitigate this issue to some degree, ANODE is still a first order ODE. This property of first order ODE limits the performance in terms of computational speed and learning of the flow. Moreover and unlike our work, those approaches are stand-alone techniques whilst our work uses a different principle -- building on ODE blocks to construct a new U-type network that can handle segmentation for complex data as in the medical domain.




\textbf{Dynamic Blocks.} \CUnet is formulated from a dynamical systems perspective. We transform the CNN boxes into second order ODEs blocks. Our blocks work under the definition of second order Neural ODEs, which read:
%
\begin{equation}
 \left\{\begin{array}{l}x^{\prime\prime}(t)=f^{(a)}\left(x(t), x^{\prime}(t), t , \theta_{f}\right) \\ x(t_0) = X_0 ,~~~ x^{\prime}(t_0)= g(x(t_0),\theta_{g}),
 \end{array}\right.
   \label{eq:important}
\end{equation}
whose velocity is described by a neural network \(f^{(a)} \) with parameters \(\theta_{f} \) and initial position given by the points of a dataset \(X_0\).





We now discuss how our dynamic block can be reliably useful and computational efficient

\begin{propb}{prop:higher_order}
Any given high-order Neural ODE can be transformed into a first-order Neural ODE. 
\end{propb}

\begin{proof}
~~Consider a first order Neural ODE:
\begin{equation}
 \left\{\begin{array}{l}x^{\prime}(t)=f^{(v)}(x(t),t,\theta_{f}) ,
 \\ x(t_0) = X_0
 \end{array}\right.
\end{equation}
where velocity is described by a neural network \( f^{(v)} \) with parameters \(\theta_{f}\). Define an \(m^{th}\) order neural ODEs as:
\begin{equation}
 \left\{\begin{array}{l} x^{m}(t)=f^{(a)}(x(t),x'(t),...,t,\theta_{f}) ,
 \\ x(t_0) = X_0,
 \\\vdots
 \\x^{m-1}(t_0)= g(x(t_0),x'(t_0),...,\theta_{g})
 \end{array}\right.
\end{equation}
where \(m \in \mathbb{Z}^+ \), and \(f^{(a)}\) refers to a neural network with parameters \(\theta_{f}\).
\begin{equation}
    Let \; \textbf{z}(t) = \begin{bmatrix}       
           x(t)=x_{1}(t) \\
           x'(t)=x_{2}(t) \\
           \vdots \\
           x^{(m-1)}(t)=x_{m}(t)
         \end{bmatrix},
        \textbf{z}(t_0) = \begin{bmatrix}
           x(t_0) \\
           x'(t_0) \\
           \vdots \\
           x^{m-1}(t_0)
        \end{bmatrix}
\end{equation}
then the \(m^{th}\) order Neural ODE becomes:
\begin{equation}
    \textbf{z}'(t) = \begin{bmatrix}       
           x_{2}(t) \\
           x_{3}(t) \\
           \vdots \\
           f^{(a)}(x_1(t),...,x_m(t),t,\theta_{f})
         \end{bmatrix}
    =f^{(v)}(\textbf{z}(t),t,\theta_{f})
\end{equation}
We therefore can always transform an \( m^{th}\) order Neural ODE into a first order Neural ODE. 
\end{proof}

The way to demonstrate that the \( m^{th}\) order ODE is reliably useful is through well-posedness. 

\begin{theob}{theo:well-posed} 
An \( m^{th}\) Order Neural ODEs with neural network $f_{\theta}$ is well-posed (in the sense of Hadamard) if $f_{\theta}$ is Lipschitz.
\end{theob}
\begin{proof}
Using Proposition~\ref{prop:higher_order}, we show that any given high order ODE can be transformed into a system of first order ODEs. 
%
The input data and output data: \( X_k\) $\mapsto$ \( Y_k\), \( k \in K \) , where \(K\) is a linearly-ordered finite subset of \(\mathbb{N} \). Then the solution of first order Neural ODEs can be solved as an initial value problem (IVP): $y(S) = h_y\Bigl(h_x(x) + \int_{s}f_{\theta(\tau)}(\tau,x,z(\tau))d\tau \Bigl)$.
%
It is then to say that if  \(f_{\theta(s)} \) is Lipschitz, then it is well-posed.

\end{proof}

\CUnet breaks the barrier of the receptive field limitation that existing U-type networks have. In existing networks, the key factor affecting the memory cost is the storing of intermediate hidden unit activation functions. Our dynamic blocks only need a single point to reconstruct the entire trajectory by forward and backward iterations. That is, \cUnet offers constant memory cost for segmentation. We guarantee this property as our blocks use trace operation via the adjoint sensitivity method with  $\mathcal{O}(1)$  memory cost.

\begin{theob}{cor:adjoint}
Our second order Neural ODE's dynamic blocks can be solved by using the first order adjoint method.
\end{theob}
\begin{proof}
Our dynamic blocks are based on second order neural ODEs. We therefore set the order to $m=2$ in Proposition~\ref{prop:higher_order} which implies  $\textbf{z}^{\prime}(t)=f^{(v)}(\textbf{z}(t),t,\theta_{f})$.
\end{proof}

Our second order dynamic blocks can then be solved by first-order adjoint method. 
At the practical level, this means that we do not need to store the layers one-by-one in the architecture. We need any single point to reconstruct the entire trajectory by forward and backward iterations. Therefore, we offer a constant memory cost.






\subsection{Opening the ODE-solver Box} \label{subsec:opening_ode_solver}
When working with Neural ODEs an open question is -- how to choose  the best ODE solver?  In this section, we derive qualitative measures for the choice of different ODE-solvers. At the implementation level, this means that our framework can be tailor-made for various segmentation tasks.  In this section, we open the ODE-solver 'box' and derive an error analysis for better understanding on how to choose different solver for \cUnet.

\begin{prf}{theo:adjoint}
Global Error (GE): $e_n = x(t_n)-x_n$ , where $x(t_n)$ denotes the exact solution and $x_n$ denotes the numerical solution. GE is the error at final time.
\end{prf}

\begin{proper}
$e_n \propto h $. If $h$ is small, higher accuracy results can be achieved.
 Moreover, if  $z=\mathcal{O}(h^p)$, we say z is p-th order where \(|z| \leq Ch^p \) , \(C >0 \) for all \( 0 <h < h_0 \). That is, when $p$ is larger, the numerical method  converges faster, being a better method in terms of convergence rate. In order to be convergent, we need to satisfy the conditions of stability and consistency.
\end{proper}


\textit{How GE can be used?} Euler's method is the most simple yet widely used one. We take it as a good starting point for more advanced methods and use our defined GE to provide further intuition. 

\begin{theob}[]{thm:GE}
Euler's method in our dynamic blocks (based on neural ODEs) converges to
 \begin{equation}
  \begin{cases}x^{\prime}(t)=\lambda x(t)+f^{(v)}(t,\theta_{f}) ~~~  0\leq t \leq t_f
  \\ x(0) = 1~~\lambda \in \mathbb{C}
   \end{cases}
 \end{equation} 
and the GE at t \(\in [0,t_f]\) is  \(\mathcal{O}(h)\).
\end{theob}

\begin{proof}

Euler’s method for our dynamic blocks gives:
\begin{equation}
        x_{n+1}  = x_n + \lambda hx_n + hf^{(v)}(t_n,\theta_{f}) 
    = (1+\lambda h )x_n + hf^{(v)}(t_n,\theta_{f})
    \label{equ:8}
\end{equation}
Applying a Taylor expansion to the exact solution, we get $x(t_{n+1}) = x(t_n) + h(\lambda x(t_n) + f^{(v)}(t_n,\theta_{f})) + R_1(t_n)$ By subtracting (\ref{equ:8}) from this expression, we find that the GE 
$e_n = x(t_n)$ - $x_n$ satisfies

\begin{equation}
    e_{n+1} =(1+h\lambda)e_n + T_{n+1}
    \label{equ:88}
\end{equation}
where write $T_{n+1}$ instead of $R_1(tn)$ to simplify the notation.  Furthermore, since $x_0 = x(t_0)$ implies $e_0 = 0$. 
We have that (\ref{equ:88}) dictates how the GE at the next step, $e_{n+1}$, combines the Local truncation error at the current step $T_{n+1}$ with the GE inherited from earlier steps $(e_n)$. A similar equation holds for more general NODE although $\lambda$ would have to be allowed to vary from step to step
Substituting $n= 0,1,2$ into (\ref{equ:88}) and using \(e_0 = 0\), \(e_1 = T_1\) and \(e_2 = (1+h\lambda)e_1 + T_2 =(1+h\lambda)T_1+T_2 \) gives
\begin{equation} 
\begin{split}
e_n & = (1+h\lambda)^{(n-1)}T_1 +(1+h\lambda)^{n-2}+...+T_n \\
  & =  \sum_{j=1}^{n} (1+h\lambda)^{n-1}T_j
\end{split}
\label{equ:11}
\end{equation}
The remaining task is finding the upper bound for the right-hand side. Firstly, we know that $|1+h\lambda| \le 1+ h|\lambda| \le e^{h|\lambda|}$ implies $|1+h\lambda|^{n-j} \le e^{(n-j)h|\lambda} = e^{|\lambda|t_{n-j}}
    \le e^{|\lambda|t_f}$.
Since \( (n-j)h = t_{n-j} \le t_f \) for \(nh \le t_f \) and \( 0 < j \le n \).
Secondly, since \( |T_j| \le Ch^2 \) for some constant \textit{C} ( independent of $h$ or $j$),
each term in the summation on the right of (\ref{equ:11}) is bounded by \( h^{2}C
e^{|\lambda|t_f} \) and so $|e_n| \le nh^{2}Ce^{|\lambda|t_f} =ht_fCe^{|\lambda|t_f}$
(by \( nh=t_f\)). Thus, assuming \(t_f\) is finite, then \(e_n = \mathcal{O}(h) \). Therefore, Euler's method applied to our dynamic blocks converges at a first-order rate, i.e., $p=1$.

\end{proof}

Whilst Euler's method is widely used as it is simple and always convergent, it is not accurate, and limited at the practical level. The GE is only $O(h)$. We therefore  provide further intuition on the family of linear multistep method (LMMs). LMMs provide a higher order $p$ which converges faster than Euler's method. 

For example, Explicit Adams-Bashforth and Implicit Adams-Bashforth-Moulton are LMMs with a higher convergent rate.
The GE of these methods is \( O(h^2) \). From a mathematical point of view, it is known that a convergent LMM is consistent.  However, the reverse does not hold, and \textit{Zero-stability} is also required~\cite{dahlquist1956convergence}. That means that Consistent + zero-stability imply convergence for these models.
If we want a method to be convergent, we need both conditions. This is the intuition behind why high order in linear multistep methods is not possible.
Finally, we discuss the Runge–Kutta (RK) method, which is another widely used and effective method.  Euler's method and midterm method are also type of RK methods with lower order. RK methods are one-step methods composed of a number of stages. The midpoint method is a RK2, i.e., a two stage method while Euler's method is RK1, i.e., a 1 stage method. \textit{Is it possible to always find a RK method of $s$ stages (order $s$)?} This is not true for \(s > 4\). 

\textit{What method is the best then?}
RK methods are a one-step methods, and the linear multistep method is a two step method. Compared to linear multistep, RK does not have to treat the first few steps taken by a single-step integration method as special cases. Moreover, RK methods are very stable. Based on our derived error analysis along with the aforementioned factors, we found that our continuous blocks, \cUnet for segmentation, greatly benefit performance-wise when using RK4. Our analysis in this section provides an answer to the open question on the solver when working with Neural ODEs. We remark that existing works on neural ODEs lack such analysis to provide a clear guidence on the solver.
 \subsection{Continuous U-net: Greater and Noiseless}\label{subsec:greater_noiseless}
This section provides the intuition on why our \cUnet is more robust and noiseless than existing U-type architectures.

Our dynamic blocks are defined as second order Neural ODEs. Existing works have shown empirically that the properties inherent to ODEs make them more robust than CNNs in terms of robustness, e.g.,~\cite{yan2019robustness}. The key idea to show such robustness comes from the ODEs integral curves as follows, which is detailed next.

\begin{theob}{theo:robustness}
ODE integral curves do not intersect (Coddington \& Levinson, 1955).
If $z_1(t)$ and $z_2(t)$ are two ODE solution with different initial value of the same function, then $z_1(t) \neq z_2(t)$ for all  
$t \in[0,\infty)$. The proof follows~\cite{coddington1955theory}.
\end{theob}

\textit{Why is} Theorem~\ref{theo:robustness} \textit{relevant?} Since ODE integral curves do not intersect, we can prove that some range always bounds the output of our dynamic blocks whilst the output of CNNs is not bounded by any range. 
This proof leads to Neural ODEs being more robust than CNNs. In particular, for our dynamic blocks based on second order ODEs, we can further support the robustness as follows: If we take a derivative on a function, we require the function to be smooth. A significant difference between the first-order ODE and second-order ODE is that the latter is at least twice continuously differentiable. However, a first-order ODE requires once continuously differentiability yielding to our dynamic blocks being smoother than first-order ODEs -- and therefore less sensitive to noise than existing techniques.

Secondly, existing U-type networks can only learn smooth homeomorphisms, which is one of the modelling disadvantages. Our blocks solve this problem by providing extra dimensions through the second order design. Moreover, our dynamic blocks are a physics-based model that better captures the  nature of segmentation. We also remark that our second-order dynamic blocks are  parameter efficient as no parameters are required on 
$x^{\prime}(t_0)= g(x(t_0),\theta_{g})$
That's why second order Neural ODEs is most robust compared to first order Neural ODEs.


\section{Experiments}
In this section, we detail the experiments that we conducted to validate our \cUnet model.


\subsection{Data Description \& Evaluation Protocol}
We expensively evaluate our \cUnet using six medical imaging datasets. They are highly heterogenous covering a wide range of medical data and significantly varying in terms of image sizes, fidelity of segmentation masks and dataset sizes. An overview of the datasets used and their properties can be found in Table \ref{table:datasets}.

\begin{table}[t!] 
\caption{Characteristics of the datasets used in our experiments.}
\centering
\resizebox{0.47\textwidth}{!}{
\begin{tabular}{ccccc}
\hline
\rowcolor[HTML]{EFEFEF} 
\textsc{Dataset} &  \# Samples  & \# Train & \# Test & Image size \\ \hline
GlaS Challenge~\cite{sirinukunwattana2017gland} & 165 & 85 & 80 & 352x352 \\
STARE~\cite{hoover2000locating} & 20 & 16 & 4 & 512x512\\
Kvasir-SEG~\cite{jha2020kvasir} & 1000 & 800 & 200 & 256x256\\
Data Science Bowl~\cite{caicedo2019nucleus} & 841 & 707 & 134 & 256x256\\
ISIC Challenge~\cite{gutman2016skin}& 1279 & 900 & 379 & 512x512\\
Breast Ultrasound Images~\cite{al2020dataset} & 647 & 518 & 129 & 256x256\\
 \hline
\end{tabular}
}
\label{table:datasets}
\end{table}

\begin{figure}
    \centering
    \includegraphics[width=1\linewidth]{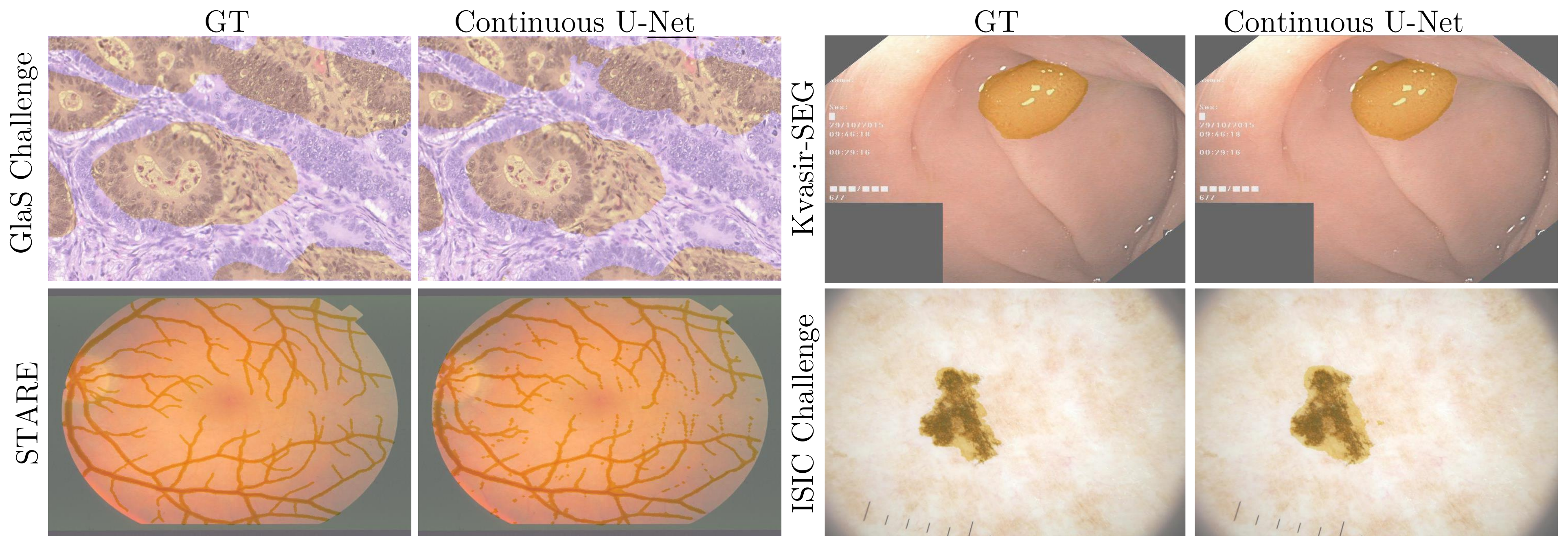}
    \caption{Visual segmentation results from our proposed \cUnet.   Extended results comparison against all techniques can be found in the supplementary material.  
    }
    \label{fig:visual_results}
\end{figure}

\begin{table*}[t!]
\centering
\caption{Comparison in terms of Dice score, accuracy and average Hausdorff distance for existing U-Net blocks and our dynamic blocks. We denote our \cUnet with \textcolor{blue}{\FiveStar}.  The best results are highlighted in green colour.
}
    \begin{subtable}{\linewidth}
\centering
\resizebox{0.95\textwidth}{!}{
\begin{tabular}{cccccccccccccc}
\hline
\rowcolor[HTML]{EFEFEF} 
\cellcolor[HTML]{FFFFFF} & \multicolumn{1}{c|}{\cellcolor[HTML]{EFEFEF}} & \multicolumn{3}{c|}{\cellcolor[HTML]{EFEFEF}GlaS Challenge} & \multicolumn{3}{c|}{\cellcolor[HTML]{EFEFEF}STARE Dataset} & \multicolumn{3}{c|}{\cellcolor[HTML]{EFEFEF}Kvasir-SEG Dataset} & \multicolumn{3}{c}{\cellcolor[HTML]{EFEFEF}DS Bowl Dataset} \\ \cline{3-14} 
\rowcolor[HTML]{FFFFFF} 
\multirow{-2}{*}{\cellcolor[HTML]{FFFFFF}} & \multicolumn{1}{c|}{\multirow{-2}{*}{\cellcolor[HTML]{EFEFEF}\begin{tabular}[c]{@{}c@{}}\textcolor{blue}{Block} \\ \textcolor{blue}{Type}\end{tabular}}} & Dice$\uparrow$ & Acc$\uparrow$ & \multicolumn{1}{c|}{\cellcolor[HTML]{FFFFFF}AHD$\downarrow$} & Dice$\uparrow$ & Acc$\uparrow$ & \multicolumn{1}{c|}{\cellcolor[HTML]{FFFFFF}AHD$\downarrow$} & Dice$\uparrow$ & Acc$\uparrow$ & \multicolumn{1}{c|}{\cellcolor[HTML]{FFFFFF}AHD$\downarrow$} & Dice$\uparrow$ & Acc$\uparrow$ & \multicolumn{1}{c}{\cellcolor[HTML]{FFFFFF}AHD$\downarrow$} \\ \hline
\rowcolor[HTML]{FFFFFF} 
\cellcolor[HTML]{FFFFFF} & \multicolumn{1}{c|}{\cellcolor[HTML]{FFFFFF}PLN} & 0.7616 & 0.7824 & \multicolumn{1}{c|}{\cellcolor[HTML]{FFFFFF}13.27} & 0.7828 & 0.9462 & \multicolumn{1}{c|}{\cellcolor[HTML]{FFFFFF}8.79} & 0.4597 & 0.8595 & \multicolumn{1}{c|}{\cellcolor[HTML]{FFFFFF}3.96} & 0.9156 & 0.9681 & \multicolumn{1}{c}{\cellcolor[HTML]{FFFFFF}3.68} \\
\cellcolor[HTML]{FFFFFF} & \multicolumn{1}{c|}{\cellcolor[HTML]{FFFFFF}RSE} & 0.7893 & 0.8100 & \multicolumn{1}{c|}{\cellcolor[HTML]{FFFFFF}12.15} & 0.7731 & 0.9460 & \multicolumn{1}{c|}{\cellcolor[HTML]{FFFFFF}8.74} & 0.6418 & 0.8754 & \multicolumn{1}{c|}{\cellcolor[HTML]{FFFFFF}4.57} & 0.9131 & 0.9687 & \multicolumn{1}{c}{\cellcolor[HTML]{FFFFFF}3.67} \\
\rowcolor[HTML]{FFFFFF} 
\cellcolor[HTML]{FFFFFF} & \multicolumn{1}{c|}{\cellcolor[HTML]{FFFFFF}DSE} & 0.7264 & 0.7516 & \multicolumn{1}{c|}{\cellcolor[HTML]{FFFFFF}13.93} & 0.7358 & 0.9413 & \multicolumn{1}{c|}{\cellcolor[HTML]{FFFFFF}8.67} & 0.6071 & 0.8649 & \multicolumn{1}{c|}{\cellcolor[HTML]{FFFFFF}4.58} & 0.9181 & 0.9690 & \multicolumn{1}{c}{\cellcolor[HTML]{FFFFFF}3.63} \\
\rowcolor[HTML]{FFFFFF} 
\cellcolor[HTML]{FFFFFF} & \multicolumn{1}{c|}{\cellcolor[HTML]{FFFFFF}INC} & 0.7962 & 0.8192 & \multicolumn{1}{c|}{\cellcolor[HTML]{FFFFFF}11.58} & 0.7750 & 0.9505 & \multicolumn{1}{c|}{\cellcolor[HTML]{FFFFFF}8.33} & 0.6856 & 0.8825 & \multicolumn{1}{c|}{\cellcolor[HTML]{FFFFFF}4.30} & 0.9210 & 0.9724 & \multicolumn{1}{c}{\cellcolor[HTML]{FFFFFF}3.41} \\
\cellcolor[HTML]{FFFFFF} & \multicolumn{1}{c|}{\cellcolor[HTML]{FFFFFF}FO} & 0.8037 & 0.8270 & \multicolumn{1}{c|}{\cellcolor[HTML]{FFFFFF}10.47} & 0.5005 & 0.9241 & \multicolumn{1}{c|}{\cellcolor[HTML]{FFFFFF}6.28} & 0.7465 & 0.9080 & \multicolumn{1}{c|}{\cellcolor[HTML]{FFFFFF}3.58} & 0.8953 & 0.9625
 & \multicolumn{1}{c}{\cellcolor[HTML]{FFFFFF}4.00} \\
\rowcolor[HTML]{FFFFFF} 
\multirow{-5}{*}{\cellcolor[HTML]{FFFFFF}\rotatebox[origin=c]{90}{ U-Net }} & \multicolumn{1}{c|}{\cellcolor[HTML]{FFFFFF}PSP} & 0.5523 & 0.5699 & \multicolumn{1}{c|}{\cellcolor[HTML]{FFFFFF}14.37} & 0.4805 & 0.9251 & \multicolumn{1}{c|}{\cellcolor[HTML]{FFFFFF}\cellcolor[HTML]{D9FFD9}5.77} & 0.5883 & 0.8672 & \multicolumn{1}{c|}{\cellcolor[HTML]{FFFFFF}4.36} & 0.9060 & 0.9655 & \multicolumn{1}{c}{\cellcolor[HTML]{FFFFFF}3.74} \\ \hline
\rowcolor[HTML]{FFFFFF} 
\begin{tabular}[c]{@{}c@{}}\textcolor{blue}{\FiveStar}\end{tabular} & \multicolumn{1}{c|}{\cellcolor[HTML]{FFFFFF}DB} & \cellcolor[HTML]{D9FFD9}0.8469 & \cellcolor[HTML]{D9FFD9}0.8675 & \multicolumn{1}{c|}{\cellcolor[HTML]{FFFFFF}\cellcolor[HTML]{D9FFD9}9.31} & \cellcolor[HTML]{D9FFD9}0.8378 & \cellcolor[HTML]{D9FFD9}0.9568 & \multicolumn{1}{c|}{\cellcolor[HTML]{FFFFFF}8.21} & \cellcolor[HTML]{D9FFD9}0.7922 & \cellcolor[HTML]{D9FFD9}0.9243 & \multicolumn{1}{c|}{\cellcolor[HTML]{FFFFFF}\cellcolor[HTML]{D9FFD9}3.32} & \cellcolor[HTML]{D9FFD9}0.9335 & \cellcolor[HTML]{D9FFD9}0.9745 & \multicolumn{1}{c}{\cellcolor[HTML]{FFFFFF}\cellcolor[HTML]{D9FFD9}3.29} \\ \hline
\multicolumn{1}{l}{} & \multicolumn{1}{l}{} & \multicolumn{1}{l}{} & \multicolumn{1}{l}{} & \multicolumn{1}{l}{} & \multicolumn{1}{l}{} & \multicolumn{1}{l}{} & \multicolumn{1}{l}{} & \multicolumn{1}{l}{} & \multicolumn{1}{l}{} & \multicolumn{1}{l}{} & \multicolumn{1}{l}{} & \multicolumn{1}{l}{} & \multicolumn{1}{l}{}
\end{tabular}
}
    \end{subtable}\par
    \begin{subtable}{\linewidth}
\centering
\resizebox{0.53\textwidth}{!}{
\begin{tabular}{cc|ccc|ccc}
\hline
 &  & \multicolumn{3}{c|}{\cellcolor[HTML]{EFEFEF}BUSI Dataset} & \multicolumn{3}{c}{\cellcolor[HTML]{EFEFEF}ISIC Challenge} \\
\multirow{-2}{*}{} & \multirow{-2}{*}{\begin{tabular}[c]{@{}c@{}}\textcolor{blue}{Block}\\ \textcolor{blue}{Type}\end{tabular}} & Dice$\uparrow$ & Acc$\uparrow$ & AHD$\downarrow$ & Dice $\uparrow$ & Acc$\uparrow$ & AHD$\downarrow$ \\ \hline
 & PLN & 0.4724 & 0.9012 & 2.56 & 0.8232 & 0.8764 & 6.76 \\
 & RSE & 0.7179 & 0.9218 & 2.63 & 0.8501 & 0.9003 & 6.00 \\
 & DSE & 0.7095 & 0.9196 & 2.61 & 0.8376 & 0.8907 & 6.32 \\
 & INC & 0.7434 & 0.9243 & 2.71 & 0.8570 & 0.9050 & 5.83 \\
 & FO & 0.7893 & 0.9338 & 2.44 & 0.9087 & \cellcolor[HTML]{FFFFFF}\cellcolor[HTML]{D9FFD9}0.9452 & \cellcolor[HTML]{FFFFFF}\cellcolor[HTML]{D9FFD9}4.45 \\
\multirow{-5}{*}{\rotatebox[origin=c]{90}{ U-Net }} & PSP & 0.6098 & 0.9197 & 2.46 & 0.8179 & 0.8798 & 6.67 \\ \hline
\textcolor{blue}{\FiveStar} & DB & \cellcolor[HTML]{FFFFFF}\cellcolor[HTML]{D9FFD9}0.8090 & \cellcolor[HTML]{FFFFFF}\cellcolor[HTML]{D9FFD9}0.9447 & \cellcolor[HTML]{FFFFFF}\cellcolor[HTML]{D9FFD9}2.15 & \cellcolor[HTML]{FFFFFF}\cellcolor[HTML]{D9FFD9}0.9094 & 0.9433 & 4.46 \\ \hline
\end{tabular}
}
    \end{subtable}
    \label{tab:blocks_performance}
\end{table*}

\begin{figure*}[t!]
    \centering
    \includegraphics[width=\textwidth]{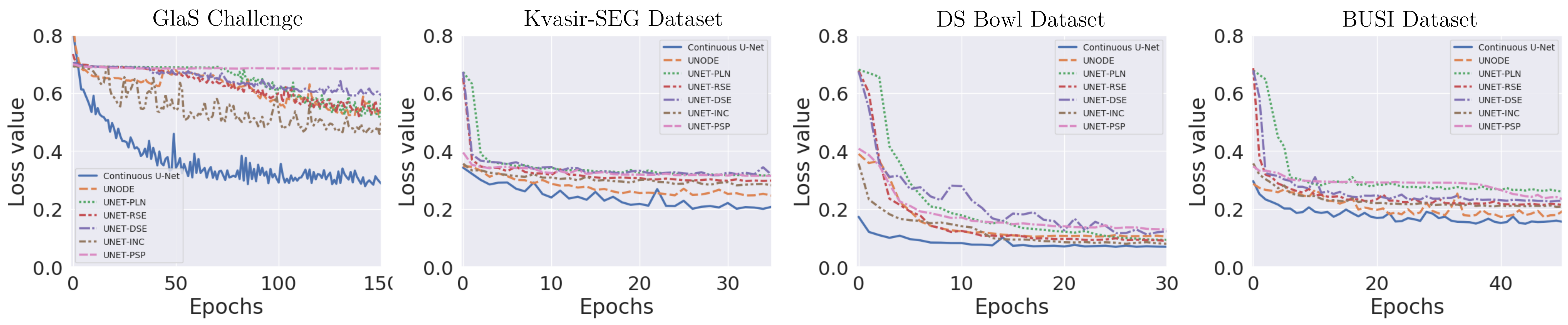}
    \caption{Convergence plots for different types of U-Net blocks. The behaviour of our dynamic blocks is denoted in blue colour, where only a few iterations are needed to converge whilst existing blocks need far longer iterations.
    }
    \label{fig:convergence}
\end{figure*}

\begin{table*}[h!]
\centering
\caption{Performance comparison, in terms of Dice score, on a range of noise level 
We denote our \cUnet with \textcolor{blue}{\FiveStar}.  The best results are highlighted in green colour. Refer to the supplementary material for extended results.
}
\resizebox{0.9\textwidth}{!}{
\begin{tabular}{cccccccccccccc}
\hline
\multicolumn{14}{c}{\cellcolor[HTML]{EFEFEF}\textsc{Noise Level Experiments}} \\ \hline
\rowcolor[HTML]{EFEFEF} 
\cellcolor[HTML]{FFFFFF} & \multicolumn{1}{c|}{\cellcolor[HTML]{FFFFFF}} & \multicolumn{4}{c|}{\cellcolor[HTML]{EFEFEF}BUSI Dataset} & \multicolumn{4}{c|}{\cellcolor[HTML]{EFEFEF}Kvasir-Seg Dataset} & \multicolumn{4}{c}{\cellcolor[HTML]{EFEFEF}ISIC Challenge} \\ \cline{3-14} 
\rowcolor[HTML]{FFFFFF} 
\multirow{-2}{*}{\cellcolor[HTML]{FFFFFF}} & \multicolumn{1}{c|}{\multirow{-2}{*}{\cellcolor[HTML]{FFFFFF}\begin{tabular}[c]{@{}c@{}}\textcolor{blue}{Block}\\ \textcolor{blue}{Type}\end{tabular}}} & 0.0 & 0.2 & 0.4 & \multicolumn{1}{c|}{\cellcolor[HTML]{FFFFFF}0.5} & 0.0 & 0.2 & 0.4 & \multicolumn{1}{c|}{\cellcolor[HTML]{FFFFFF}0.5} & 0.0 & 0.2 & 0.4 & 0.5 \\ \hline
\rowcolor[HTML]{FFFFFF} 
\cellcolor[HTML]{FFFFFF} & \multicolumn{1}{c|}{\cellcolor[HTML]{FFFFFF}PLN} & 0.4724 & 0.4723 & 0.4722 & \multicolumn{1}{c|}{\cellcolor[HTML]{FFFFFF}0.4721} & 0.4597 & 0.4597 & 0.4597 & \multicolumn{1}{c|}{\cellcolor[HTML]{FFFFFF}0.4597} & 0.8232 & 0.4064 & 0.4064 & 0.4064 \\
\rowcolor[HTML]{FFFFFF} 
\cellcolor[HTML]{FFFFFF} & \multicolumn{1}{c|}{\cellcolor[HTML]{FFFFFF}RSE} & 0.7179 & 0.2373 & 0.2130 & \multicolumn{1}{c|}{\cellcolor[HTML]{FFFFFF}0.2094} & 0.6418 & 0.4597 & 0.4597 & \multicolumn{1}{c|}{\cellcolor[HTML]{FFFFFF}0.4597} & 0.8501 & 0.4064 & 0.4064 & 0.4064 \\
\rowcolor[HTML]{FFFFFF} 
\cellcolor[HTML]{FFFFFF} & \multicolumn{1}{c|}{\cellcolor[HTML]{FFFFFF}DSE} & 0.7095 & 0.1944 & 0.2321 & \multicolumn{1}{c|}{\cellcolor[HTML]{FFFFFF}0.2423} & 0.6071 & 0.4597 & 0.4597 & \multicolumn{1}{c|}{\cellcolor[HTML]{FFFFFF}0.4597} & 0.8376 & 0.4774 & 0.4067 & 0.4064 \\
\rowcolor[HTML]{FFFFFF} 
\cellcolor[HTML]{FFFFFF} & \multicolumn{1}{c|}{\cellcolor[HTML]{FFFFFF}INC} & 0.7434 & 0.1930 & 0.1769 & \multicolumn{1}{c|}{\cellcolor[HTML]{FFFFFF}0.1883} & 0.6856 & 0.4597 & 0.4597 & \multicolumn{1}{c|}{\cellcolor[HTML]{FFFFFF}0.4597} & 0.8570 & 0.4064 & 0.4064 & 0.4064 \\
\rowcolor[HTML]{FFFFFF} 
\multirow{-5}{*}{\cellcolor[HTML]{FFFFFF}\rotatebox[origin=c]{90}{ U-Net }} & \multicolumn{1}{c|}{\cellcolor[HTML]{FFFFFF}PSP} & 0.6098 & 0.5899 & 0.5368 & \multicolumn{1}{c|}{\cellcolor[HTML]{FFFFFF}0.5255} & 0.5883 & 0.4668 & 0.4597 & \multicolumn{1}{c|}{\cellcolor[HTML]{FFFFFF}0.4597} & 0.8179 & 0.3979 & 0.3851 & 0.3847 \\ \hline
\rowcolor[HTML]{FFFFFF} 
\textcolor{blue}{\FiveStar} & \multicolumn{1}{c|}{\cellcolor[HTML]{FFFFFF}DB} & \cellcolor[HTML]{D9FFD9}0.8090 & \cellcolor[HTML]{D9FFD9}0.6881 & \cellcolor[HTML]{D9FFD9}0.5743 & \multicolumn{1}{c|}{\cellcolor[HTML]{FFFFFF}\cellcolor[HTML]{D9FFD9} 0.5569} & \cellcolor[HTML]{D9FFD9}0.7922 & \cellcolor[HTML]{D9FFD9}0.6417 & \cellcolor[HTML]{D9FFD9}0.6164 & \multicolumn{1}{c|}{\cellcolor[HTML]{FFFFFF}\cellcolor[HTML]{D9FFD9}0.5972} & \cellcolor[HTML]{D9FFD9}0.9094 & \cellcolor[HTML]{D9FFD9}0.7055 & \cellcolor[HTML]{D9FFD9}0.6910 & \cellcolor[HTML]{D9FFD9}0.6746 \\ \hline
\end{tabular}
}
\label{tab:noise}
\end{table*}

\textbf{Evaluation Protocol.} Following a standard protocol 
 in medical image segmentation, we evaluate the performance of the proposed \cUnet and existing techniques using three metrics: the Dice score, accuracy and averaged Hausdorff distance. For a fair comparison, we use a shared code-base for all experiments. More precisely, we set a learning rate of $1\times 10^{-3}$,
a step-based learning rate scheduler with a step size of $1$ and a gamma value of $0.999$. We use a fourth-order Runge-Kutta (RK4) solver, 
a batch size of $16$ and train all networks for $500$ epochs.

\begin{table*}
\caption{Ablation study on different types of ODE solvers on four medical image segmentation datasets. Fourth-order Runge-Kutta (RK4) consistently outperforms Euler, Explicit Adams-Bashforth (AB) and Implicit Adams-Bashforth-Moulton (ABM) in Dice score, accuracy and average Hausdorff distance. The best results are highligthed in green colour.
Refer to the supplementary material for extended results.}
\centering
\resizebox{0.95\textwidth}{!}{
\begin{tabular}{c|ccc|ccc|ccc|ccc}
\hline
 & \multicolumn{3}{c|}{\cellcolor[HTML]{EFEFEF}STARE Dataset} & \multicolumn{3}{c|}{\cellcolor[HTML]{EFEFEF}DS Bowl Dataset} & \multicolumn{3}{c|}{\cellcolor[HTML]{EFEFEF}Kvasir-SEG Dataset} & \multicolumn{3}{c}{\cellcolor[HTML]{EFEFEF}ISIC Challenge} \\
\multirow{-2}{*}{\begin{tabular}[c]{@{}c@{}}\textcolor{blue}{Solver}\\ \textcolor{blue}{Type}\end{tabular}} & Dice$\uparrow$ & Acc$\uparrow$ & AHD$\downarrow$ & Dice $\uparrow$ & Acc$\uparrow$ & AHD$\downarrow$ & Dice $\uparrow$ & Acc$\uparrow$ & AHD$\downarrow$ & Dice $\uparrow$ & Acc$\uparrow$ & AHD$\downarrow$ \\ \hline
 Euler & 0.7697 & 0.9468 & 8.57 & 0.8911 & 0.9539 & 4.37 & 0.5450 & 0.8596 &4.82 & 0.8487 & 0.9003 & 6.91 \\
 AB & 0.8066 & 0.9517 & 8.49 & 0.9252 & 0.9726 & 3.39 & 0.7850 & 0.9185 & 3.46 & 0.9051 & 0.9398 & 4.68 \\
 ABM & 0.8015 & 0.9506 & 8.50 & 0.9246 & 0.9730 & 3.37 & 0.7714 & 0.9158 & 3.50 & 0.9032 & 0.9374 & 4.78 \\
 RK4 &\cellcolor[HTML]{FFFFFF}\cellcolor[HTML]{D9FFD9} 0.8378 & \cellcolor[HTML]{FFFFFF}\cellcolor[HTML]{D9FFD9}0.9568 & \cellcolor[HTML]{FFFFFF}\cellcolor[HTML]{D9FFD9}8.2149 & \cellcolor[HTML]{FFFFFF}\cellcolor[HTML]{D9FFD9}0.9335 & \cellcolor[HTML]{FFFFFF}\cellcolor[HTML]{D9FFD9}0.9745 & \cellcolor[HTML]{FFFFFF}\cellcolor[HTML]{D9FFD9}3.29 & \cellcolor[HTML]{FFFFFF}\cellcolor[HTML]{D9FFD9}0.7922 & \cellcolor[HTML]{FFFFFF}\cellcolor[HTML]{D9FFD9}0.9243 & \cellcolor[HTML]{FFFFFF}\cellcolor[HTML]{D9FFD9}3.32 & \cellcolor[HTML]{FFFFFF}\cellcolor[HTML]{D9FFD9}0.9094 & \cellcolor[HTML]{FFFFFF}\cellcolor[HTML]{D9FFD9}0.9433 & \cellcolor[HTML]{FFFFFF}\cellcolor[HTML]{D9FFD9}4.46 \\ \hline
\end{tabular}
}
\label{tab:solver_analysis}
\end{table*}

\begin{table*}
\caption{Performance comparison of \cUnet against  state-of-the-art U-Net architectures with additional mechanisms. We denote our \cUnet with \textcolor{blue}{\FiveStar}. Extended results with additional three datasets can be found in the supplementary material. }
\centering
\resizebox{0.9\textwidth}{!}{
\begin{tabular}{c
>{\columncolor[HTML]{FFFFFF}}c |
>{\columncolor[HTML]{FFFFFF}}c 
>{\columncolor[HTML]{FFFFFF}}c 
>{\columncolor[HTML]{FFFFFF}}c |
>{\columncolor[HTML]{FFFFFF}}c 
>{\columncolor[HTML]{FFFFFF}}c 
>{\columncolor[HTML]{FFFFFF}}c |
>{\columncolor[HTML]{FFFFFF}}c 
>{\columncolor[HTML]{FFFFFF}}c 
>{\columncolor[HTML]{FFFFFF}}c }
\hline
\cellcolor[HTML]{EFEFEF} & \cellcolor[HTML]{EFEFEF} & \multicolumn{3}{c|}{\cellcolor[HTML]{EFEFEF}GlaS Challenge} & \multicolumn{3}{c|}{\cellcolor[HTML]{EFEFEF}STARE Dataset} & \multicolumn{3}{c}{\cellcolor[HTML]{EFEFEF}Kvasir-SEG Dataset} \\ \cline{3-11} 
\multirow{-2}{*}{\cellcolor[HTML]{EFEFEF}\textsc{Technique}} & \multirow{-2}{*}{\cellcolor[HTML]{EFEFEF}\textsc{Mechanism}} & Dice$\uparrow$ & Acc$\uparrow$ & AHD$\downarrow$ & \cellcolor[HTML]{FFFFFF}Dice$\uparrow$ & \cellcolor[HTML]{FFFFFF}Acc$\uparrow$ & \cellcolor[HTML]{FFFFFF}AHD$\downarrow$ & \cellcolor[HTML]{FFFFFF}Dice$\uparrow$ & \cellcolor[HTML]{FFFFFF}Acc$\uparrow$ & \cellcolor[HTML]{FFFFFF}AHD$\downarrow$ \\ \hline
\cellcolor[HTML]{FFFFFF}\begin{tabular}[c]{@{}c@{}}Attn. U-Net\end{tabular} & \begin{tabular}[c]{@{}c@{}}Attn. Gates\end{tabular} & 0.7991 & 0.8173 & 12.10 & 0.8675 & 0.9637 & 7.63 & 0.7290 & 0.8997 & 4.35 \\
\cellcolor[HTML]{FFFFFF}DynU-Net & Heuristic-R + Opt & 0.8626 & 0.8828 & 9.15 & 0.8508 & 0.9575 & 8.25 & 0.7634 & 0.9101 & 4.00 \\
\cellcolor[HTML]{FFFFFF}U2Net & \begin{tabular}[c]{@{}c@{}}Nested U-Nets\end{tabular} & 0.8465 & 0.8658 & 9.18 & 0.7845 & 0.9539 & 7.84 & 0.7950 & 0.9171 & 3.23 \\
UNeXt & Tokenised MLP & 0.8911 & 0.9040 & 7.72 & 0.7866 & 0.9482 & 8.59 & 0.7879 & 0.9260 & 3.08 \\ 
\cellcolor[HTML]{FFFFFF}TransUnet & Transformers & 0.8984 & 0.9145 & 7.11 & 0.8565 & 0.9607 & 7.79 & 0.8749 & 0.9429 & 2.70 \\ \hline
\cellcolor[HTML]{FFFFFF}\textcolor{blue}{\FiveStar} & \xmark & 0.8469 & 0.8675 & 9.31 & 0.8378 & 0.9568 & 8.21 & 0.7922 & 0.9243 & 3.32 \\ \hline
\end{tabular}
}
\label{tab:u_net_variants_performance}
\end{table*}

\subsection{Results \& Discussion}
In this section, we present all experimental results and visualisation conducted to validate our approach.

\smallskip
\fcircle[fill=ocra]{2pt} \textbf{Comparison to Other Existing U-Type Blocks.} We underline that our work is a stand-alone continuous network, where the highlight is the new dynamic block. Therefore, our main comparison is based on different types of U-Net blocks.
More precisely, our comparison includes plain convolutional blocks (PLN), residual blocks (RSE), dense blocks (DSE), inception blocks (INC), pyramid pooling blocks (PSP), first order ODE blocks (FO) and our dynamic blocks (DB).  We start by reporting the global results in Table~\ref{tab:blocks_performance}. The displayed numbers show all metrics. Looking at the results more closely, we observe that we achieve significant improvement over most SOTA techniques, most notable on the GlaS and the Kvasir-SEG challenge datasets. There are only three cases where the FO block performs similarly well than our dynamic blocks. However, a closer look at the FO block shows general difficulties to get reliable performance for the remaining datasets -- as for example in the STARE dataset. In contrast, our dynamic blocks report a stable and high performance across all datasets and metrics. Additionally, Figure \ref{fig:visual_results} highlights visual results for our approach on different datasets.


\fcircle[fill=ocra]{2pt} \textbf{\CUnet: Faster \& Greater}
To further support our theoretical findings, we evaluate the convergence of our \cUnet against the blocks of other existing techniques.  
In Figure~\ref{fig:convergence} we display convergence plots for challenging datasets. We observe that our theory agrees with the experiments; as our dynamic blocks offer an extra dimension, by the second-order Neural ODEs modelling, which yields to require fewer iterations for the solution than other existing U-type blocks. In a closer look, we observe  that our dynamic blocks only need a small number of epochs to converge whilst the other ones do not converge even with a higher number of epochs.


\fcircle[fill=ocra]{2pt} \textbf{\CUnet: Noiseless} 
In Subsection \ref{subsec:greater_noiseless}, we proved that our \cUnet is less sensitive to noise than other CNN-based U-Net architectures.  
We use pretrained models, adding zero mean Gaussian noise with varying standard deviations ($0.2, 0.4, 0.5$)  and compute the Dice score for each of them. Table~\ref{tab:noise} reports the results and highlights the robustness of our approach. \CUnet consistently achieves the highest Dice scores for all noise levels. Additionally, comparing our \cUnet approach to the second best performing version -- the U-Net with inception blocks (INC) -- the difference becomes even more obvious. For all datasets, adding even a small amount of noise (standard deviation of $0.2$) to the INC model, the Dice score drops massively, e.g., from $0.7434$ to $0.1930$ on the GlaS challenge dataset. Our approach, however, is able to deal much better with additive noise, leading to a difference in Dice score of only $0.1209$. A similar pattern can be observed for the other two datasets.  

\fcircle[fill=ocra]{2pt} \textbf{\CUnet: Opening the ODE-Solver Box} Subsection \ref{subsec:opening_ode_solver} sheds light onto the theoretical error analysis for different types of ODE solvers.  We compare Euler's method, Explicit Adams-Bashforth (AB), Implicit Adams-Bashforth-Moulton (ABM) and a fourth-order Runge-Kutta (RK4) 
Table \ref{tab:solver_analysis} reports the performance for all methods and datasets.
Our derived qualitative measures agree with the empirical results, 
the Runge-Kutta method outperforms Euler, Adam-Bashforth and Adam-Bashforth-Moulton for each of the datasets in all metrics and is thus suggested for segmentation tasks. 

\fcircle[fill=ocra]{2pt} \textbf{\CUnet vs. other U-type with Additional Mechanisms.}
We underline that our main contribution in this paper is to create a continuous network with our dynamic blocks. Whilst a direct comparison with architectures using additional mechanisms is unfair, we ran a set of experiments on current SOTA techniques with additional mechanisms including attention gates, transformers or tokenised MLPs. Firstly, we seek to demonstrate that our network stands alone without any additional mechanism. Secondly, to open the door to a new research line to design continuous U-type networks with additional mechanisms-- which is far from being trivial.  Table \ref{tab:u_net_variants_performance} provides the results for these experiments. \CUnet is able to outperform at least two methods per dataset (GlaS challenge, STARE dataset, DS Bowl dataset, ISIC challenge),   three for the Kvasir-SEG dataset and even four on the BUSI dataset.  \vspace{-0.15cm}

\section{Conclusion}
\CUnet is a continuous network modelled by our dynamic blocks using second order neural ODEs. We show that our approach outperforms existing U-Net blocks on six  benchmarking datasets, and readily competes  or even outperforms famous U-Net architectures with additional mechanisms like attention. 
Our findings open the door to a new research line on continuous U-type networks by introducing the to the best of our knowledge first U-type architecture that provides underpinning theory.

\section*{Acknowledgements}
CWC acknowledges support from Department of Mathematics, College of Science , CityU and HKASR reaching out award.
AIAR acknowledges support from CMIH and CCIMI, University of Cambridge.
CBS acknowledges support from the Philip Leverhulme Prize, the Royal Society Wolfson Fellowship, the EPSRC advanced career fellowship EP/V029428/1, EPSRC grants EP/S026045/1 and EP/T003553/1, EP/N014588/1, EP/T017961/1, the Wellcome Innovator Awards 215733/Z/19/Z and 221633/Z/20/Z, the European Union Horizon 2020 research and innovation programme under the Marie Skodowska-Curie grant agreement No. 777826 NoMADS, the Cantab Capital Institute for the Mathematics of Information and the Alan Turing Institute.


{\small
\bibliographystyle{ieee_fullname}
\bibliography{egbib}}

\clearpage

\clearpage

\clearpage
\begin{table}[h!]
\centering
\begin{tabular}{c}
\begin{tabular}[c]{@{}c@{}} \large  \textsc{Supplemental Material} \\ \\  \Large \textbf{Continuous U-Net:} \\\Large \textbf{Faster, Greater and Noiseless }\end{tabular}
\end{tabular}
\end{table}

\bigskip
\section*{A. Outline}
This document extends the practicalities and results presented in the maing paper. This is structured as follows. 

\begin{itemize}
    \item \textbf{Supplementary Convergence \& Noise Results:} We provide further results comparisons of our \cUnet against other existing U-type networks. We also give further experimental results on noisy data. 
    \item \textbf{Supplementary Solver Results:} We provide further results on the ablation study of our model under different type of ODE solvers and its effect.
    \item \textbf{Supplementary Performance Comparison \& Training Scheme.} We give further numerical comparisons for different type of blocks along with other type of U-Nets with different mechanisms. In the interest of clarity and completness, we give an explicit definition in the training setting of our model. 
\end{itemize}


\subsection{Supplementary Convergence \& Noise Results }
\begin{figure}[h!]
	\centering
	\includegraphics[width=\linewidth]{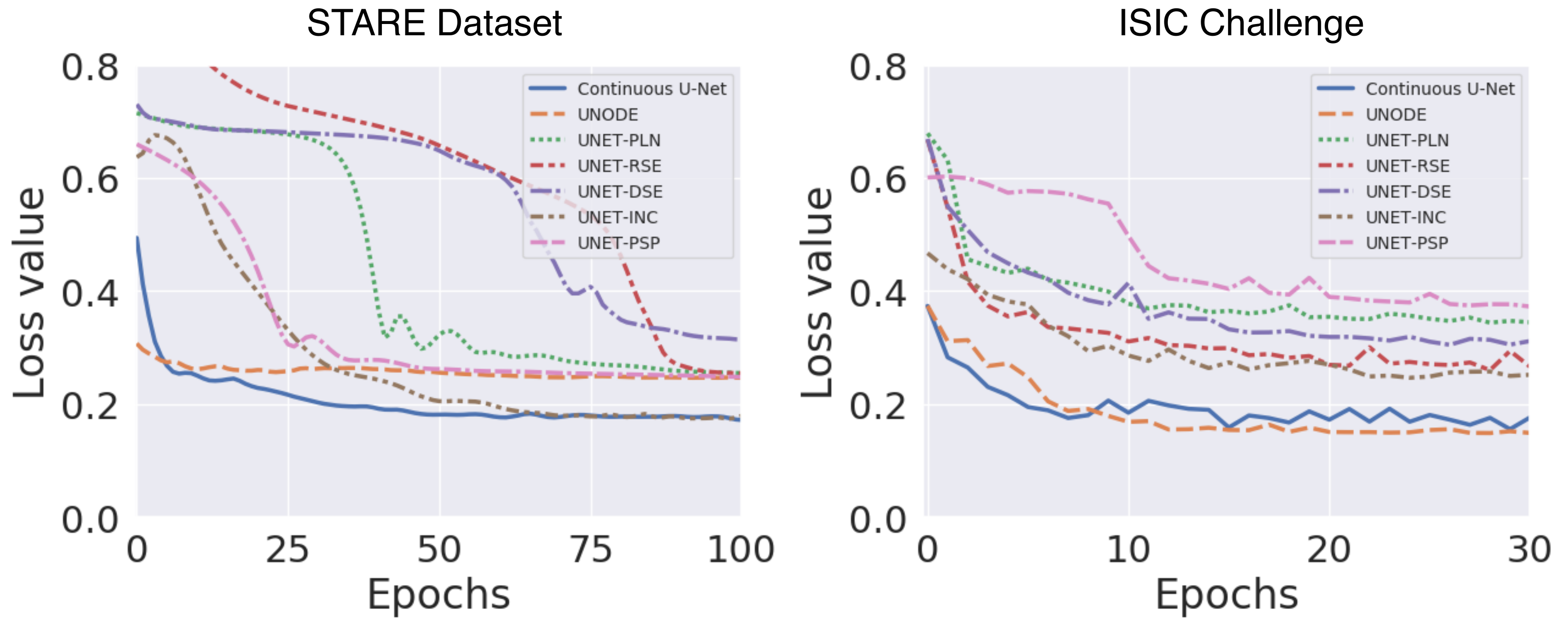}
	\caption{Convergence plots for different types of U-Net blocks on remaining two datasets. The behaviour of our dynamic blocks is denoted in blue colour, where only a few iterations are needed to converge whilst existing blocks need far more iterations.}
 \label{fig:supp_convergence}
\end{figure}

Figure \ref{fig:supp_convergence} provides convergence plots for different types of U-Net blocks for the STARE dataset and the ISIC challenge. Our \cUnet is denoted in blue colour. In a closer look, we observe that our model only need a few iterations to converge whilst existing blocks need far more iterations.

We also extend our noise results from the main paper using  the GlaS challenge, STARE dataset and DS Bowl dataset. Our experients are conducted by adding zero mean Gaussian noise with increasing standard deviation. 
The results are in  Table \ref{tab:supp_noise}.  They shows that except for the DS Bowl dataset, our approach overall performs the best in terms of Dice metric. This is the only dataset where our approach achieves lower  scores when adding noise, while it outperforms the other block types for the other datasets, we show that it is less sensitive to noise than the other block types in most of the cases.

\subsection{Supplementary Solver Results}
In addition to the experimental results displayed in the main paper, we presented some further experiments for the different type of solvers and its effect in our model, displayed in  Table~\ref{tab:supp_solver_analysis}. The results show that the fourth order Runge-Kutta solver performs the best for the BUCSI dataset in terms of accuracy and Average Hausdorff distance. For the particular case of the GlaS dataset, however, the implicit Adam-Bashforth-Moulton solver outperforms the RK4 one. Our theroretical insights reported in the main paper follows the empirical findings showing that RK4 is theoretically and empirically is the best choice.

\begin{table}[h!]
\caption{Ablation study on different types of ODE solvers on the two remaining medical image segmentation datasets. The best results are highligthed in green colour.}
\centering
\resizebox{0.48\textwidth}{!}{
\begin{tabular}{c|ccc|ccc}
\hline
 & \multicolumn{3}{c|}{\cellcolor[HTML]{EFEFEF}GlaS Challenge} & \multicolumn{3}{c}{\cellcolor[HTML]{EFEFEF}BUSI Dataset} \\
\multirow{-2}{*}{\begin{tabular}[c]{@{}c@{}}\textcolor{blue}{Solver}\\ \textcolor{blue}{Type}\end{tabular}} & Dice$\uparrow$ & Acc$\uparrow$ & AHD$\downarrow$ & Dice $\uparrow$ & Acc$\uparrow$ & AHD$\downarrow$ \\ \hline
 Euler & 0.7141 & 0.7374 & 14.01 & 0.6612 & 0.8957 & 3.36\\
 AB & 0.8455 & 0.8668 & 9.32 & 0.8127 & 0.9419 & 2.28  \\
 ABM & \cellcolor[HTML]{FFFFFF}\cellcolor[HTML]{D9FFD9}0.8521 & \cellcolor[HTML]{FFFFFF}\cellcolor[HTML]{D9FFD9}0.8696 & \cellcolor[HTML]{FFFFFF}\cellcolor[HTML]{D9FFD9}9.27
 & \cellcolor[HTML]{FFFFFF}\cellcolor[HTML]{D9FFD9}0.8203 & 0.9409
 & 2.34 \\
 RK4 & 0.8469 & 0.8675 & 9.31 & 0.8090 & \cellcolor[HTML]{FFFFFF}\cellcolor[HTML]{D9FFD9}0.9447
 & \cellcolor[HTML]{FFFFFF}\cellcolor[HTML]{D9FFD9}2.15 \\ \hline
\end{tabular}
}
\label{tab:supp_solver_analysis}
\end{table}

\subsection{Supplementary Performance Comparison \& Training Scheme}
We provide further numerical comparison of our model against other existing U-type Networks using additional mechanisms. The results are displayed in Table~\ref{tab:u_net_variants_performance} using the  DS Bowl dataset, BUSI dataset and ISIC challenge. As for the other datasets, our apporach is able to perform as good as most of the state of the art approaches while even outperforming some of them, even though despite we do not use any additional mechanisms.
Finally and for sake of clarity, we provide the training scheme we for our experiments, which is summarised in Table \ref{tab:supp_overview_training_settings}. We remark that we use the same setting for  all experiments for a fair comparison.

\begin{table*}[t!]
\centering
\caption{Performance comparison, in terms of Dice score, on a range of noise levels.
We denote our \cUnet with \textcolor{blue}{\FiveStar}. The best results are highlighted in green colour.
}
\resizebox{0.9\textwidth}{!}{
\begin{tabular}{cccccccccccccc}
\hline
\multicolumn{14}{c}{\cellcolor[HTML]{EFEFEF}\textsc{Noise Level Experiments}} \\ \hline
\rowcolor[HTML]{EFEFEF} 
\cellcolor[HTML]{FFFFFF} & \multicolumn{1}{c|}{\cellcolor[HTML]{FFFFFF}} & \multicolumn{4}{c|}{\cellcolor[HTML]{EFEFEF}GlaS Challenge} & \multicolumn{4}{c|}{\cellcolor[HTML]{EFEFEF}STARE Dataset} & \multicolumn{4}{c}{\cellcolor[HTML]{EFEFEF}DS Bowl Dataset} \\ \cline{3-14} 
\rowcolor[HTML]{FFFFFF} 
\multirow{-2}{*}{\cellcolor[HTML]{FFFFFF}} & \multicolumn{1}{c|}{\multirow{-2}{*}{\cellcolor[HTML]{FFFFFF}\begin{tabular}[c]{@{}c@{}}\textcolor{blue}{Block}\\ \textcolor{blue}{Type}\end{tabular}}} & 0.0 & 0.2 & 0.4 & \multicolumn{1}{c|}{\cellcolor[HTML]{FFFFFF}0.5} & 0.0 & 0.2 & 0.4 & \multicolumn{1}{c|}{\cellcolor[HTML]{FFFFFF}0.5} & 0.0 & 0.2 & 0.4 & 0.5 \\ \hline
\rowcolor[HTML]{FFFFFF} 
\cellcolor[HTML]{FFFFFF} & \multicolumn{1}{c|}{\cellcolor[HTML]{FFFFFF}PLN} & 0.7616 & 0.5898
 & 0.4209
 & \multicolumn{1}{c|}{\cellcolor[HTML]{FFFFFF}0.3977
} &  0.7828
 & \cellcolor[HTML]{D9FFD9}0.5554
 & 0.4929
 & \multicolumn{1}{c|}{\cellcolor[HTML]{FFFFFF}0.4840
} & 0.9156
 & \cellcolor[HTML]{D9FFD9}0.7635
 & \cellcolor[HTML]{D9FFD9}0.6563
 & \cellcolor[HTML]{D9FFD9}0.6253
 \\
\rowcolor[HTML]{FFFFFF} 
\cellcolor[HTML]{FFFFFF} & \multicolumn{1}{c|}{\cellcolor[HTML]{FFFFFF}RSE} & 0.7893
 &  0.4813
& 0.5133
 & \multicolumn{1}{c|}{\cellcolor[HTML]{FFFFFF}0.5117
} & 0.7731
 & 0.4805
 & 0.4805
 & \multicolumn{1}{c|}{\cellcolor[HTML]{FFFFFF}0.4805
} & 0.9131
 &  0.5015
 & 0.4892
 & 0.4863
 \\
\rowcolor[HTML]{FFFFFF} 
\cellcolor[HTML]{FFFFFF} & \multicolumn{1}{c|}{\cellcolor[HTML]{FFFFFF}DSE} & 0.7264
 & 0.4564
 & 0.3713
 & \multicolumn{1}{c|}{\cellcolor[HTML]{FFFFFF}0.3613
} & 0.7358
 & 0.4805
 & 0.4805
 & \multicolumn{1}{c|}{\cellcolor[HTML]{FFFFFF}0.4805
} & 0.9181
 & 0.5307
 &0.4842
& 0.4698
 \\
\rowcolor[HTML]{FFFFFF} 
\cellcolor[HTML]{FFFFFF} & \multicolumn{1}{c|}{\cellcolor[HTML]{FFFFFF}INC} & 0.7962
 & 0.4517
 & 0.4485
 & \multicolumn{1}{c|}{\cellcolor[HTML]{FFFFFF}0.4697
} & 0.7750
 & 0.5290
 & 0.4895
 & \multicolumn{1}{c|}{\cellcolor[HTML]{FFFFFF}0.4829
} & 0.9210
 & 0.5172
 & 0.4849
 & 0.4751
 \\
\rowcolor[HTML]{FFFFFF} 
\multirow{-5}{*}{\cellcolor[HTML]{FFFFFF}\rotatebox[origin=c]{90}{ U-Net }} & \multicolumn{1}{c|}{\cellcolor[HTML]{FFFFFF}PSP} & 0.5523
 &0.5523
& 0.5515
 & \multicolumn{1}{c|}{\cellcolor[HTML]{FFFFFF}\cellcolor[HTML]{D9FFD9}0.5469
} & 0.4805
 & 0.4805
 & 0.4806
 & \multicolumn{1}{c|}{\cellcolor[HTML]{FFFFFF}0.4805
} & 0.9060
 & 0.4895
 & 0.4690
& 0.4669
 \\ \hline
\rowcolor[HTML]{FFFFFF} 
\textcolor{blue}{\FiveStar} & \multicolumn{1}{c|}{\cellcolor[HTML]{FFFFFF}DB} & \cellcolor[HTML]{D9FFD9}0.8469
 & \cellcolor[HTML]{D9FFD9} 0.8063
& \cellcolor[HTML]{D9FFD9} 0.5882
& \multicolumn{1}{c|}{\cellcolor[HTML]{FFFFFF}0.4922
 } & \cellcolor[HTML]{D9FFD9} 0.8378
& 0.5145
 & \cellcolor[HTML]{D9FFD9} 0.5034
& \multicolumn{1}{c|}{\cellcolor[HTML]{FFFFFF}\cellcolor[HTML]{D9FFD9}0.5017
} & \cellcolor[HTML]{D9FFD9} 0.9335
& 0.4939
 & 0.4678
 & 0.4644
 \\ \hline
\end{tabular}
}
\label{tab:supp_noise}
\end{table*}

\begin{table*}[t!]
\caption{Performance results for state of the art U-Net architectures with additional mechanisms. We mark our \cUnet with \textcolor{blue}{\FiveStar}. }
    
\centering
\resizebox{0.95\textwidth}{!}{
\begin{tabular}{
>{\columncolor[HTML]{FFFFFF}}c 
>{\columncolor[HTML]{FFFFFF}}c |
>{\columncolor[HTML]{FFFFFF}}l 
>{\columncolor[HTML]{FFFFFF}}l 
>{\columncolor[HTML]{FFFFFF}}l |
>{\columncolor[HTML]{FFFFFF}}l 
>{\columncolor[HTML]{FFFFFF}}l 
>{\columncolor[HTML]{FFFFFF}}l |
>{\columncolor[HTML]{FFFFFF}}l 
>{\columncolor[HTML]{FFFFFF}}l 
>{\columncolor[HTML]{FFFFFF}}l }
\hline
\cellcolor[HTML]{EFEFEF} & \cellcolor[HTML]{EFEFEF} & \multicolumn{3}{c|}{\cellcolor[HTML]{EFEFEF}DS Bowl Dataset} & \multicolumn{3}{c|}{\cellcolor[HTML]{EFEFEF}BUSI Dataset} & \multicolumn{3}{c}{\cellcolor[HTML]{EFEFEF}ISIC Challenge} \\ \cline{3-11} 
\multirow{-2}{*}{\cellcolor[HTML]{EFEFEF}\textsc{Technique}} & \multirow{-2}{*}{\cellcolor[HTML]{EFEFEF}\textsc{Mechanism}} & \multicolumn{1}{c}{\cellcolor[HTML]{FFFFFF}Dice$\uparrow$} & \multicolumn{1}{c}{\cellcolor[HTML]{FFFFFF}Acc$\uparrow$} & \multicolumn{1}{c|}{\cellcolor[HTML]{FFFFFF}AHD$\downarrow$} & \multicolumn{1}{c}{\cellcolor[HTML]{FFFFFF}Dice$\uparrow$} & \multicolumn{1}{c}{\cellcolor[HTML]{FFFFFF}Acc$\uparrow$} & \multicolumn{1}{c|}{\cellcolor[HTML]{FFFFFF}AHD$\downarrow$} & \multicolumn{1}{c}{\cellcolor[HTML]{FFFFFF}Dice$\uparrow$} & \multicolumn{1}{c}{\cellcolor[HTML]{FFFFFF}Acc$\uparrow$} & \multicolumn{1}{c}{\cellcolor[HTML]{FFFFFF}AHD$\downarrow$} \\ \hline
\begin{tabular}[c]{@{}c@{}}Attn. U-Net\end{tabular} & \begin{tabular}[c]{@{}c@{}} Attn. Gates\end{tabular} & 0.9388 & 0.9760 & 3.16 & 0.7484 & 0.9208 & 3.01 & 0.8987 & 0.9331 & 4.98 \\
DynU-Net & Heuristic-R + Opt & 0.9446 & 0.9771 & 3.07 & 0.7716 & 0.9270 & 2.83 & 0.9105 & 0.9408 & 4.48 \\
U2Net & \begin{tabular}[c]{@{}c@{}}Nested U-Nets\end{tabular} & 0.4776 & 0.8723 & 5.97 & 0.5096 & 0.8696 & 4.12 & 0.8310 & 0.8863 & 7.83 \\
UNeXt & Tokenised MLP & 0.6671 & 0.8943 & 5.54 & 0.8064 & 0.9398 & 2.26 & 0.9202 & 0.9482 & 4.01 \\ 
TransUnet & Transformers & 0.9351 & 0.9760 & 3.17 & 0.8507 & 0.9497 & 2.00 & 0.9251 & 0.9529 & 3.74 \\ \hline
\cellcolor[HTML]{FFFFFF}\textcolor{blue}{\FiveStar} & \xmark  & 0.9335 & 0.9745 & 3.29 & 0.8090 & 0.9447 & 2.15 & 0.9094 & 0.9433 & 4.46 \\ \hline
\end{tabular}
}

    \label{tab:u_net_variants_performance}
\end{table*}

\begin{table*}[t!]
    \centering
    \caption{Overview of training settings for all experiments.}
    \begin{tabular}{l r}
    \toprule
      \textbf{Parameter}   &  \textbf{Value} \\ \midrule
      Loss function & Binary Cross Entropy Loss \\ 
      Optimiser & Adam\\ 
      Learning rate & $10^{-4}$\\ 
      Learning rate schedule & Multiplication of learning rate with $0.999$ every epoch\\ 
      Epochs & 500 \\ 
      Batch size & 16\\  
      Levels of U-Net architecture & 4 \\
      Number of filters per block & 3, 6, 12, 24 \\
      Tolerance (for contin. blocks only) & $10^{-3}$ \\\bottomrule
    \end{tabular}
    \label{tab:supp_overview_training_settings}
\end{table*}

\end{document}